# The 1st Data Science for Pavements Challenge

Ashkan Behzadian[1], Tanner Wambui Muturi[2], Tianjie Zhang[3], Hongmin Kim[3] Amanda Mullins[3],
Yang Lu[3], Neema Jasika Owor[4], Yaw Adu-Gyamfi[5], William Buttlar[5], Majidifard Hamed[5],
Armstrong Aboah[5], David Mensching[6], Spragg Robert[6], Matthew Corrigan[6],
Jack Youtchef[6], Dave Eshan[7]

[1] Eastern Medditerranean University, Famagusta, North Cyprus
[2] Middle East Technical University, Ankara, Turkey
[3] Boise State University, Boise Idaho, USA
[4] Eastern Makerere University, Uganda
[5] University of Missouri – Columbia, MO, USA
[6] Federal Highway Administration, DC, USA
[7] University of New Hampshire, NH, USA

*Abstract*— The Data Science for Pavement Challenge (DSPC) seeks to accelerate the research and development of automated vision systems for pavement condition monitoring and evaluation by providing a platform with benchmarked datasets and codes for teams to innovate and develop machine learning algorithms that are practice-ready for use by industry. The first edition of the competition attracted 22 teams from 8 countries. Participants were required to automatically detect and classify different types of pavement distresses present in images captured from multiple sources, and under different conditions. The competition was data-centric: teams were tasked to increase the accuracy of a predefined model architecture by utilizing various data modification methods such as cleaning, labeling and augmentation. A real-time, online evaluation system was developed to rank teams based on the F1 score. Leaderboard results showed the promise and challenges of machine for advancing automation in pavement monitoring and evaluation. This paper summarizes the solutions from the top 5 teams. These teams proposed innovations in the areas of data cleaning, annotation, augmentation, and detection parameter tuning. The F1 score for the top-ranked team was approximately 0.9. The paper concludes with a review of different experiments that worked well for the current challenge and those that did not yield any significant improvement in model accuracy.

## I. INTRODUCTION

Pavement distress evaluation systems are an essential component of any strategy developed for prioritizing pavement for maintenance and rehabilitation. The most prevalent types of pavement surface distress are longitudinal, transverse, diagonal, block, and alligator cracks. Typically, field engineers detect pavement distress through visual inspection. With the rise of AI in recent years, different fields of engineering and science are embracing the use of machine learning-enabled object recognition systems [1]–[3]. AI models are, however, data hungry; the lack of massive data to train machine learning algorithms, the requirement for expensive, high-end computing resources, and the lack of interpretability of how these models arrive at their decisions is impeding the adoption of AI technologies across many transportation agencies. In response to these challenges, we have launched the first edition of the data science for pavements challenge (DSPC). In collaboration with the Federal Highway Administration, State Departments of Transportation and academia, the challenge seeks to accelerate the research and development of automated vision systems for pavement condition monitoring and evaluation. The competition provides access to massive amounts of labeled data to feed machine learning- based algorithms. Organizers developed a platform to share benchmarked datasets; providing an environment that enables teams to innovate and address real-world pavement engineering problems, as well as evaluated their algorithms against datasets and metrics.

There are two fundamental components of Artificial Intelligence: Data and Model. Although data appears to be a central object of machine learning, most data competitions and challenges have been centered around iterating on things such as algorithm design, model architecture design and feature engineering. One of many examples of such competitions is the Global Road Damage Detection Challenge (GRRDC) which ended in 2020 in the US [4]. The competition aimed to detect and classify different pavement distress on a large dataset of front-view images captured using smartphones and acquired from different countries. The competition was model-centric and resulted in the development of many state-of-the-art solutions and models as found in [5], [6]. However, the top-ranked team had an F-1 score of 0.676 after numerous model iterations including ensemble learning, hyperparameter tuning and optimizations. The level of accuracy attained is not sufficient to attract broad adoption of the technology by the pavement evaluation industry. This outcome also proves that focusing on model iterations alone may only lead to marginal improvements in accuracies. To overcome this challenge, there is a need to also iterate through the data; investigating data manipulation strategies that could results in significant improvements in model accuracies: Hence, the introduction of DSPC.

In model-centric AI, the data is often treated as exogenous from the machine learning development process. However, in recent years, AI researchers are quickly realizing that focusing on data iterations is more crucial to the successful and rapid development of accurate and transferable models [7]–[9]. The DSPC is designed to follow data-centric AI principles: where one augments, cleans, and annotates datasets to improve the accuracy of a predefined model. In this approach, the model is kept constant, while one iteratively improves the data quality. Compared to GRRDC, the current competition is data-centric: aiming to increase the accuracy of a predefined model architecture by utilizing various data modification methods such as cleaning, labeling and augmentation. The perspective of images provided in DSPC are top-down compared to the dashcam-view used by GRRDC. The remainder of this paper summarizes the 2022 DSPS Challenge preparation and results. In the following sections, we describe the challenge task and data preparation, benchmark annotation process, evaluation methodology, analysis of submitted results from top-ranked teams, and a discussion of insights and future trends.

## II. CHALLENGE TASK

DSPSDC required participants to automatically detect and classify different types of pavement distresses present in images captured from multiple sources, and under different conditions. The DSPSDC followed a data-centric model instead of the traditional model-centric approaches [4]. With the data-centric model, participants are required to systematically change/enhance datasets provided using various data cleaning, annotation, augmentation strategies to improve the accuracy of a predefined model architecture. Although the hyperparameters of the pre-defined model can be altered, participants were restricted to use the same model architecture. The organizers selected a light-weight version of Yolov5 as the network architecture for this competition. Using this model and following data-centric modeling principles, participants were tasked to generate bounding boxes and labels for each damage instance present in the testing database. Scripts were provided to generate a JSON file from model predictions, which were submitted to the competition website for scoring. The two main general rules of the competition were as follows:

- Data – Participants could only use the data provided. The use of external datasets were prohibited.
- Algorithms and Model – The algorithm and model (including pre-trained models) used by participants was predefined. Teams were encouraged to leverage hyper parameter tuning, data cleaning and augmentation strategies to improve the performance of the pre-defined model.

## III. DATASET

The dataset used for the competition were top-down view of pavement images collected from three cities including Kansas City, Jefferson City and Columbia Missouri. The images captured varying distress type, extent, and severity under diverse conditions. The DSPS Data Set consisted of the following data sources:

- ARAN Data: About 70 percent of the dataset were obtained from ARAN vehicles. Top-down images are obtained directly from high-speed cameras. The ARAN vehicles use an external lighting source to eliminate uneven illumination and shadows from trees and vehicles. However, artifacts such as oil stains were visible on images. The resolution of images produced from the ARAN vans was 1080 x 960.
- Google Street View: This dataset was captured using Google's API for extracting images from Street View. The API extracts images given a location, heading, pitch, and zoom level. A pitch of 270 degrees was sufficient to provide top-down views similar to the images obtained from the ARAN. The dataset had artifacts such as shadows of trees, vehicles, and oil stains. The resolution of images extracted from Google Street View was 640 x 640.

## IV. ANNOTATION AND CLASS LABELING

Bounding box annotation of distresses were carried out using CVAT's online annotation tool. The tool enabled collaborative annotation of images by volunteers and pavement engineers. About 800 images were annotated using this tool. Annotators assigned one of seven labels per distress. The distress labels included Alligator, Block, Transverse, Patching, Sealing, Longitudinal, and Manhole.

Seven student volunteers were used to annotate the training and test datasets. To ensure consistency in labeling, each student was assigned only one of the seven class labels. The quality of annotations from students were judged by four experienced pavement engineers and researchers to ensure that participating teams had access to high quality data to train and test their models. The annotation effort of each student volunteer was evaluated by two judges. The judges corrected annotations by changing class labels and resizing the bounding box around wrongly annotated distresses. Figure 1 shows example images and their corresponding labels (each color representing a different class).

The datasets were split into three sections (training set 1, training set 2 and test). The first batch of training data was released about 7 weeks before the competition deadline. It contained all pavement images and corresponding labels. The second training dataset was released 2 weeks after the first. Participants were provided with pavement images with no labels. Each participant was expected to generate their own annotations using the annotation tool provided. This s e t up enabled the organizers to evaluate the influence of labeling on the performance of deep models. The test data was releases about 3 weeks to the submission deadline.

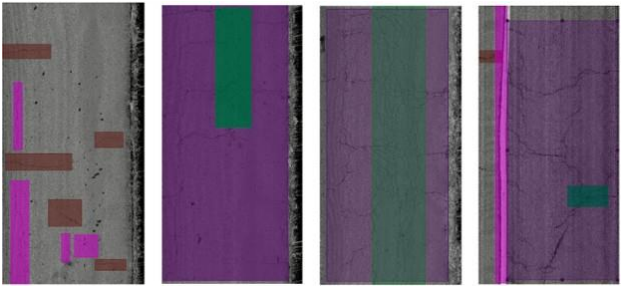

Fig. 1: Sample Annotations of longitudinal, transverse, block and alligator cracking.

Similar to the second training data, it consisted of images with no annotations. Participants were instructed to generate labels automatically using their develop models only.

The DSPS datasets were made available to participants in two popular formats: Pascal VOC and DarkNet. Scripts were also made available to convert between other popular annotation formats, generate their own training, validation and testing splits.

## V. PARTICIPATING TEAMS

22 teams worldwide registered for the challenge. 11 teams were shortlisted based on the average F1-score achieved. Table 1 shows a leaderboard of the top 11 teams, their ranks, and scores. For the final round of evaluations, the top 5

TABLE I: Leaderboard

| Team | Rank | Accuracy |
|---|---|---|
| Ashkan | 1 | 0.8952915179330274 |
| neema | 2 | 0.76004828002414 |
| pavers | 3 | 0.7128753180661578 |
| Taniac | 4 | 0.6584463625154131 |
| Mistletoe | 5 | 0.6332599118942731 |
| $Na_S$ | 6 | 0.6209453197405004 |
| Sygna | 7 | 0.6138392857142858 |
| LexIT | 8 | 0.6123280692817117 |
| DDK | 9 | 0.5765765765765766 |
| KHS | 10 | 0.5647461494580719 |
| Frances | 11 | 0.539119804400978 |

## VI. EVALUATION

Each submission was evaluated as a function of the mean of per-class F-1 scores. The F1 score, measures accuracy using the statistics precision and recall. Precision is the ratio of true positives to all predicted positives. Recall is the ratio of true positives to all actual positives. The F1 metric weights recall and precision equally, and a good retrieval algorithm will maximize both precision and recall simultaneously. Thus, moderately good performance on both will be favored over excellent performance on one and poor performance on the other. The F-1 score is designated as equation 1.

TABLE II: Influence of annotation quality of model – heatmap of f1score, confusion matrix influence.

| Team Number and Name | Proposed Solutions |
|---|---|
| 1 – Ashkan_bhz | - Training and test-time augmentation including bounding box safe-crop, image inversion and mosaic.<br>- Trial and error tuning of detection parameters such as non-max-suppression IOU thresholds and detection confidence. |
| 2 - Pavers | - Semi-supervised data annotation<br>- Training data augmentation with mosaic, blurring, scaling, flipping and image inversion. |
| 3 - Tanic | - Training data augmentation including horizontal and vertical flipping, random adjustment of hue and contrast.<br>- Trial and error tuning of detection parameters. |
| 4 - Mistletoe | - Training data augmentation with GANS.<br>- Training data augmentation with mean normalization, and histogram equalization, |
| 5 - Neema | - Augmented annotation<br>- Detection parameter tuning.<br>- Training data augmentation with mosaic, flipping, and image blurring. |

$$F1 = \frac{2 * precision * recall}{precision + recall} \quad (1)$$

$$precision = \frac{TruePositives}{TruePositives + FalsePositives} \quad (2)$$

$$recall = \frac{TruePositives}{TruePositives + FalseNegatives} \quad (3)$$

1) The area of overlap between the predicted bounding box and ground truth bounding box divided by the area of union of the boxes (also called IOU) exceeds 50 percent.
2) The predicted label matches the actual label.

A REST API was developed to allow for online evaluation of results submitted by participants. The API allowed unlimited submissions from teams, enabling them to improve the results over time. Evaluation scores of each submission were stored in a real-time database and displayed on a leaderboard (as shown in Table 1) where teams were able to see their F-1 scores and rank.

## VII. SUBMISSION RESULTS, RANKINGS AND EXPERIMENTS

This section summarizes results from the top 5 teams of the competition. Table 2 captures details of the methods used by the top-ranked teams including hyperparameter tuning, data cleaning, labeling and augmentation. The average F1- score for these top teams was more than 70 percent. The winning teams followed four key strategies in tackling the pavement distress recognition task: (1) image denoising and enhancement (2) auto labeling, (3) data augmentation, and

(4) test-time augmentation. Image Denoising and Enhancement: Team 4 leveraged image enhancement techniques prior to model development. Pavement surface images in general are affected by pulse interference. Therefore, a Gaussian filter is used to denoise the images. Furthermore, histogram equalization is utilized to enhance defects, thus making them more recognizable. Considering defects are only identified by different colors compared to the entire pavement surface, an algorithm which minus the mean pixel value from the whole image is utilized to remove the noise and make the defects clearer. Figure 2 below shows the influence of these pre-processing algorithms on model accuracy. Both gaussian filtering and histogram equalization decreased the models' ac- curacy. There were no significant difference between scores for mean normalization and using the raw image. Most deep convolutional neural network architectures implement these pre-processing algorithms by default, this could be the reason for the negligible effect of applying these image denoising and enhancement techniques before training. Annotation and Auto Labeling: The competition released two datasets for model training. The first batch of training data was labeled by organizers. The teams were expected to label the second batch of training data. Teams 2 and 5 leveraged an auto- labeling technique to reduce the time needed to annotate the second training dataset. They automatically generated labels for the second batch of training data using a model pretrained on the first batch of training data. The teams were able to focus on correcting only false labels instead of annotating all distress in each image manually. Addition- ally, teams were able to gain insight into instances where model failed and adapted subsequent techniques such as data augmentation to overcome these limitations. The annotations from teams were inspected and compared to ground truth annotations from the organizers. The accuracy of manual annotations by teams is shown in Figure 3 below. It was realized that most of the teams confused alligator and block cracks. Also, cracks that had diagonal orientations were not consistently assigned to a unique class. The influence of the manual annotation accuracy on the model's accuracy is also shown in Figure 3. For most deep learning models, the rule of thumb is that with increasing size of training data, models generalize better and therefore, its accuracy on the test data also improves. However, for some teams, model accuracy dropped or increased marginally after adding the second batch of training data which they manually annotated. It is important to note that although high annotation accuracies (> 90 percent) results in about 3 – 5 percent increase in models' accuracy, a poorly labeled datasets (75 percent accurate) only reduced models' accuracy by just 1 percent. Data Augmentation: The number of images in the test set provided was equal to the training set. The test set also had more diverse data in terms of the severity of distresses and pavement type. Data augmentation was therefore crucial to avoid overfitting – a situation in which the network memorizes the training data features instead of learning them: this results in the model's accuracy on the unseen data (testing data) being drastically lower than training data [10]. Almost all the teams used some form of data augmentation to overcome this problem. Teams 2, 3, and 4 used the following transformers:

- Mosaic: allows models to learn how to detect objects at smaller scales by combining four training images into one in certain ratios. This augmentation technique also significantly reduces the need for a large mini-batch size.
- Scaling: zooms in/out of the original image. Other geometric transformations such as rotations were not used as it could lead to confusion between transverse and longitudinal cracks.
- Flipping: Flips the images horizontal and or vertically.

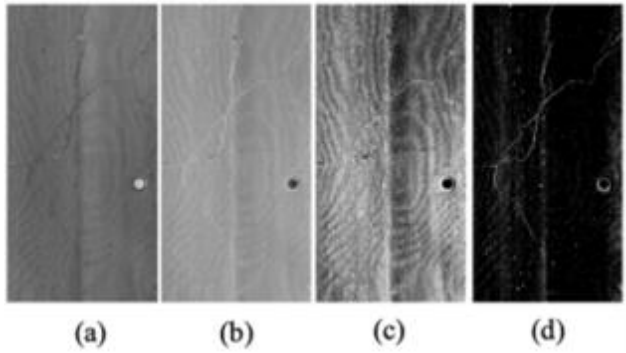 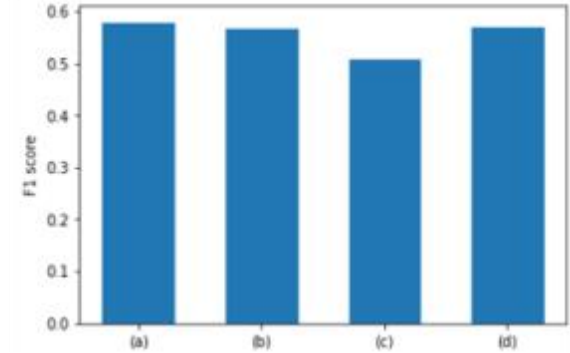

Fig. 2: Influence of different processing techniques on model accuracy: left – a). Raw image; b) Gaussian Filter; c) Histogram equalization d) Image mean normalization. Right – model accuracies given a) Raw unprocessed image b) Gaussian filtered image c) Image processed using histogram equalization d) Mean normalized image.

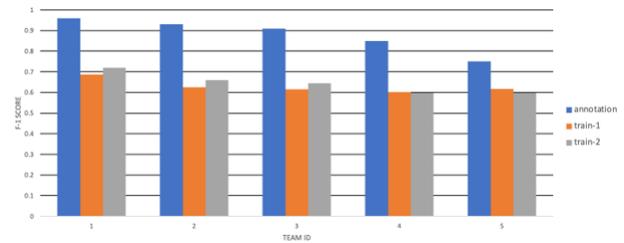

Fig. 3: Influence of annotation quality of model – heatmap of f1score, confusion matrix influence.

The first-place team used the afore-mentioned transformers in addition to the following methods:

- Bounding Box Safe Crop: this method crops the image in a way that the bounding boxes will not remove from

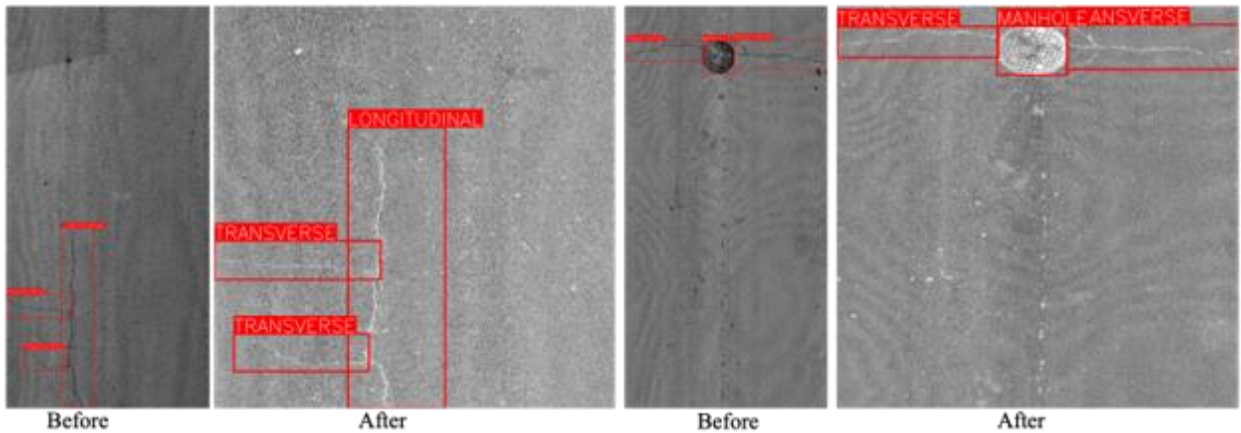

Fig. 4: Sample images before and after apply augmentation.

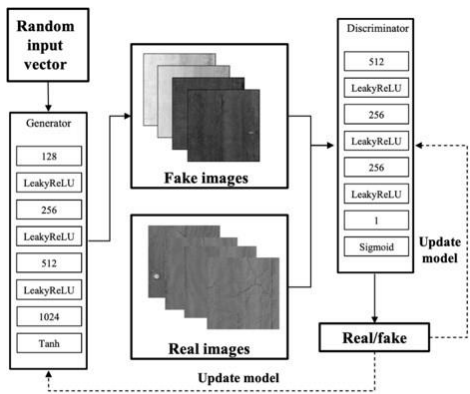

Fig. 5: Structure of GAN.

the original image.
- Image Inversion: this method inverts the image by subtracting pixel values by 225.

Team 4 explored generative adversarial networks (GANS) to generate additional, realistic pavement image dataset for training purposes. Figure 5 shows an overview of the GAN used for the competition. The generator is provided with random input vectors of the equivalent size as real images, to produce fake images. These fake images are then compared with real images in the discriminator. The outcome of the discriminator will update the model and the parameters in the generator to help it produce more realistic fake images. The loss function used in GAN is Binary Cross-Entropy and the optimizer used is Adaptive Moment Estimation.

Figure 6 shows sample pavement images that were generated with GANs and its influence on the overall model accuracy. The addition of fake images increased the f1-score by about 9 percent from 0.581 to 0.633.

Test-Time Augmentation (TTA): TTA involves using models to make predictions on multiple augmented copies of each image in the test set, and then returning an ensemble of the output predictions. TTA gives the model the best opportunity to correctly classify test images. The first-place team was the only team that leveraged TTA on the test data provided. The team applied three main transformations to the test data: flipping, image inversion and bounding box safe crop. These transportations were used to then generate 10 augmented copies generated per image. In addition to these modifications, most teams observed that adjusting the prediction confidence (CONF) and non-max-suppression (NMS) IOU thresholds dramatically influenced the result's accuracy. The non-max-suppression IOU threshold represents that the model accepts or denies a predicted bounding box. Figure 7 shows the influence of different combinations of CONF, NMS and augmentation types on the model accuracy.

The first column of the figure shows results when all the augmentation techniques discussed are used. The second and third columns shows effect of not using mosaic, scaling, and flipping augmentation. The result shows about 10 percent drop in the accuracy of the model when these augmentation techniques are not used during training. There was however no significant impact on model accuracy when they were n o t used for test time augmentation. Image inversion and bounding box safe crop had the most significant impact on the models' accuracy as shown in the last column of Figure 7. The models accuracy dropped by about 17 percent and 10 percent when this augmentation technique was removed during training and test time respectively.

VII. DISCUSSIONS

The 1st edition of the DSPS Challenge received strong participation and quality submissions from research communities around the world. Top-ranked teams had collaborations between civil engineers with domain expertise in pavement crack evaluation and computer scientists with machine learning skillset. Teams leveraged data-centric machine learning principles to significantly improved the accuracy of a predefined model without changing its architecture. In this section, we summarize the key observations from the challenge results, including both successful and failed experiments.

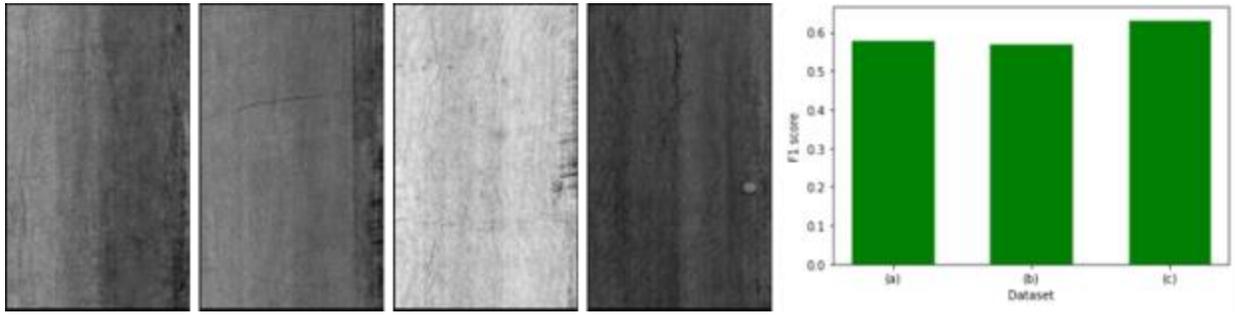

Fig. 6: Sample pavement images generated with GANs and a plot of the influence of increasing training dataset with GAN generated images – a). first batch of training data (b) after adding the second release annotated by team (c) after adding the selected fake image from GANS

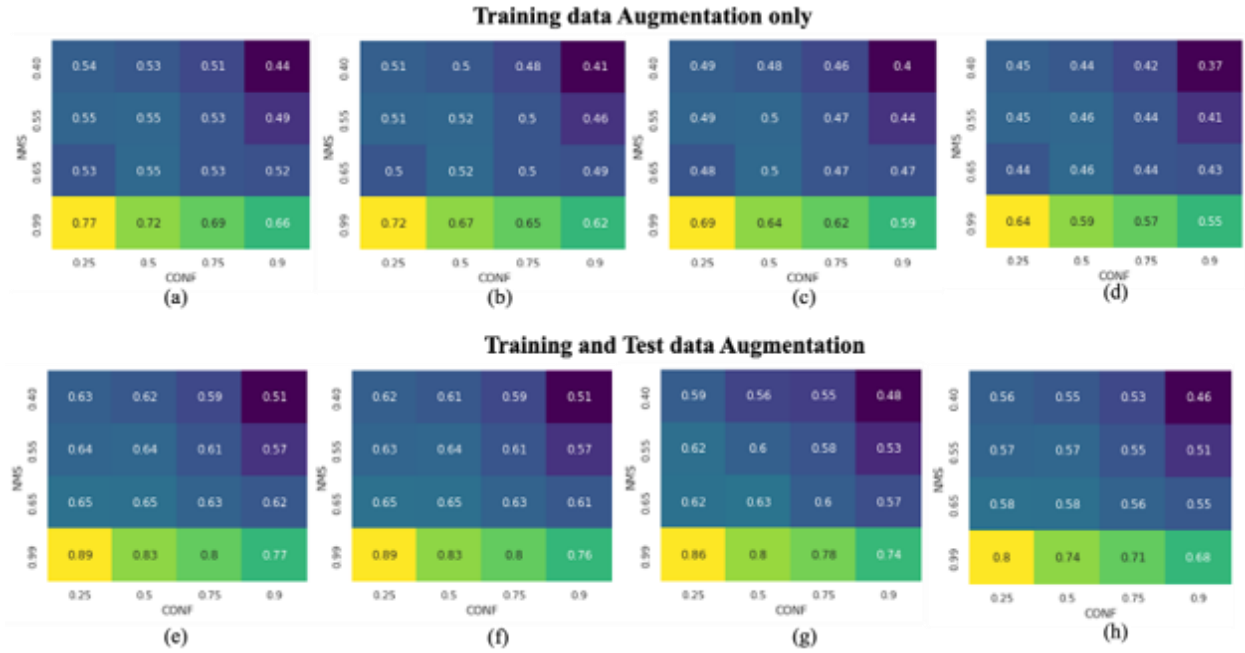

Fig. 7: Influence of augmentation type on model accuracy: first row applies augmentation to training datasets only, while the second rows show results after augmenting both training and test datasets. a and e - applies all five augmentation techniques (mosaic, scaling, flipping, safe crop and image inversion), b and f shows results after eliminating mosaic data augmentation, c and g removes scaling and flipping augmentation while d and h shows results if bounding box safe crop and image inversion are removed (no augmentation).

*A. Successful Experiments*

A successful experiment refers to any data iteration technique that resulted in significant improvement in model performance. Augmented annotation, Training data and Test-time augmentation were by far the most influential techniques for improving model accuracy. These techniques resulted in about 20 – 30 percent increase in model accuracy. Annotation: The quality of training data annotations can- not be over emphasized. Misclassifications and incorrectly positioned bounding boxes can deteriorate the models performance. The competition however realized that when training data is sufficient, the models' performance may not be significantly impacted if the labeling errors are within 5 percent. Beyond this error rate, the model's performance could plateau even after increasing training data size. Another crucial innovation was the so-called augmented or semi-supervised labeling. Semi-supervised annotation can significantly improve the speed of object class labeling for large image databases. The process involves building an intermediate model from a small sample of labeled training data and then using the developed model to automatically label the remaining training dataset. Although manual verifications and corrections will still be needed, it significantly reduces the manual labeling requirements. Teams were also able to use this technique to understand how the models' reasoning for distress classification.

Augmentation: Data augmentation must be tailored to the problem being solved. Not all data augmentation processes improve model's accuracy and generalization to unseen data. In fact, some have a degrading effect on the model output. For example, for pavement distress recognition, it is essential to avoid the augmentation method that changes the cracks' orientation, such as rotation and shear because it might confuse the model for detecting the type of cracks, specifically the transverse and longitudinal cracks. For instance, by rotating a longitudinal crack for 90 degrees, the direction of the crack could be similar to a transverse crack. On the other hand, several proper data augmentation methods improve the generalization and enhance the model's accuracy. One of the most effective methods could be cropping. As illustrated in the previous sections, the distresses in some of the pavement images occurs on a specific part of the image, and the rest of the image does not contain distress. In this case, cropping the part of the image which contains the distresses and eliminating the rest of the image could be extremely helpful for the model to distinguish the distresses among the background. Another proper method could be image inversion. By applying the image inversion method on the pavement distress images, the color of the cracks in the original images, which are mainly darker than the background, will be brighter. In this case, the model's performance would be improved by distinguishing the cracks on the pavement's surface. The following figure (Figure 8) represents the performance of the training loss before and after applying the abovementioned augmentation methods. It can be seen that, before applying the augmentation methods, the training loss has an up-trending increase after 20 epochs which obviously shows the overfitting effect. In contrast, after applying proper augmentation methods, the training loss fluctuated after 50 epochs.

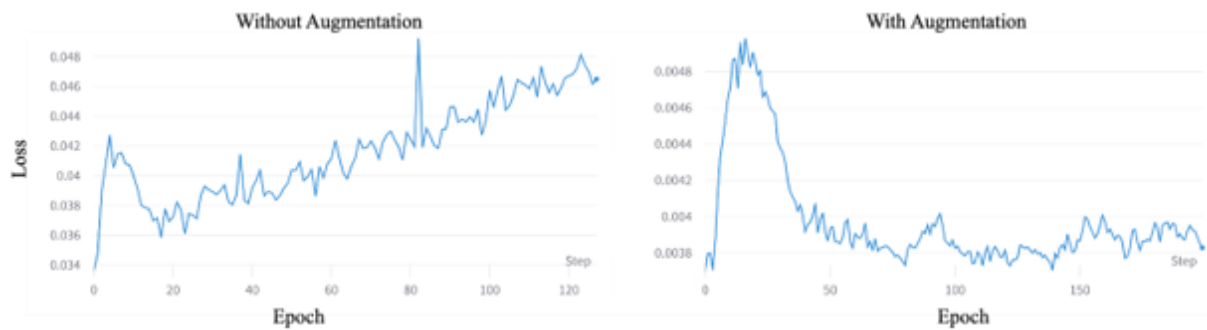

Fig. 8: Influence of data augmentation technique on training and validation loss convergence.

### B. Failed Experiments

There were a number of data manipulations and iterations and experiments which surprisingly did not result in any significant improvement in model accuracy. Data augmentation techniques which smooth or changes the general contrast of the pavement images failed to yield any significant improvement in model accuracy. These types of augmentation techniques include median blur, and random brightness contrast. Lastly, although transfer learning was used by all top-ranked teams there were no significant difference in terms of model accuracy gains for different types of transfer learning techniques such as freezing all layer weights (except the last convolutional layer), backbone, and other intermediate layers. GANS for instance, were surprisingly not as influential given that it was used to increase the number of training dataset: it yielded only about 9 percent increase in model accuracy.

### VIII. CONCLUSION AND FUTURE SCOPE

The Data Science for Pavements Challenge focuses on the application of artificial intelligence for pavement condition monitoring. Top-down views of pavement image data containing 7 distress types annotated (Alligator, Block, Transverse, Patching, Sealing, Longitudinal, and Manhole) with bounding boxes were provided to teams. The images were acquired from multiple sources in three different cities and under different conditions. Through the DSPC platform, we solicited for contributions that leveraged data-centric machine learning principles to improve the performance of state-of-the-art deep learning models. The results from top ranking teams show promise for practical and large-scale deployment of these models for road maintenance prioritization and overall, data-driven decision making. The competition showed that the type of augmentation used for the data iteration processes can have significant improvement (or degradation) on the performance of deep models. For pavement distress recognition, bounding box safe crop and image inversion where the most influential with about 10 percent increase in model accuracy. While rotations resulted in degrading effect, other techniques such as median blurring, mosaic and image flipping had marginal effects. The first Data Science for Pavements Challenge has seen strong participation from teams all over the world, with about 22 teams submitting results that significantly improved the baselines on these challenging tasks. The next edition of the DSPS will maintain the data-centric machine learning approach. The challenge will however evolve from a single task to a multi-task prediction problem. In addition to detecting and classifying distresses, participants will be tasked to quantify the extent and severity of the predicted class. The competition will also develop new metrics to test for the transferability of models developed by participants. For example, models trained on top-down view images will be tested on images captured at slightly different perspectives (e.g., front view cameras). With regards to benchmarked datasets, we will increase the number of classes to include distress from both asphalt and concrete pavements: reflective, bleeding, raveling, PCC joint, D cracking and Spalling. Competition organizers will also explore benchmarking datasets which captures both surface and sub-surface information for pavement condition monitoring. Machine learning models will then be used to predict future evolution of degradation.